\documentclass[11pt,a4paper]{article}
\usepackage[utf8]{inputenc}
\usepackage{amsmath,amssymb}
\usepackage{bm} 
\usepackage{graphicx}
\usepackage{hyperref}
\usepackage{geometry}
\usepackage{booktabs} 
\usepackage{adjustbox} 
\usepackage{caption} 
\usepackage{subcaption} 

\hypersetup{
    colorlinks=true,
    linkcolor=blue,
    filecolor=magenta,      
    urlcolor=cyan,
    citecolor=red, 
}

\title{Comparing the Moore–Penrose Pseudoinverse and Gradient Descent for Solving Linear Regression Problems: A Performance Analysis}
\author{Alex Adams \\ University of Pittsburgh \\ \texttt{asa298@pitt.edu}}
\date{May 29, 2025} 

\begin{document}
\maketitle

\begin{abstract}
This paper investigates the comparative performance of two fundamental approaches to solving linear regression problems: the closed-form Moore–Penrose pseudoinverse and the iterative gradient descent method. Linear regression is a cornerstone of predictive modeling, and the choice of solver can significantly impact efficiency and accuracy. I review and discuss the theoretical underpinnings of both methods, analyze their computational complexity, and evaluate their empirical behavior on synthetic datasets with controlled characteristics, as well as on established real-world datasets. My results delineate the conditions under which each method excels in terms of computational time, numerical stability, and predictive accuracy. This work aims to provide practical guidance for researchers and practitioners in machine learning when selecting between direct, exact solutions and iterative, approximate solutions for linear regression tasks.
\end{abstract}

\section{Introduction}
\begin{figure}[h!]
\centering
 \includegraphics[width=0.8\textwidth]{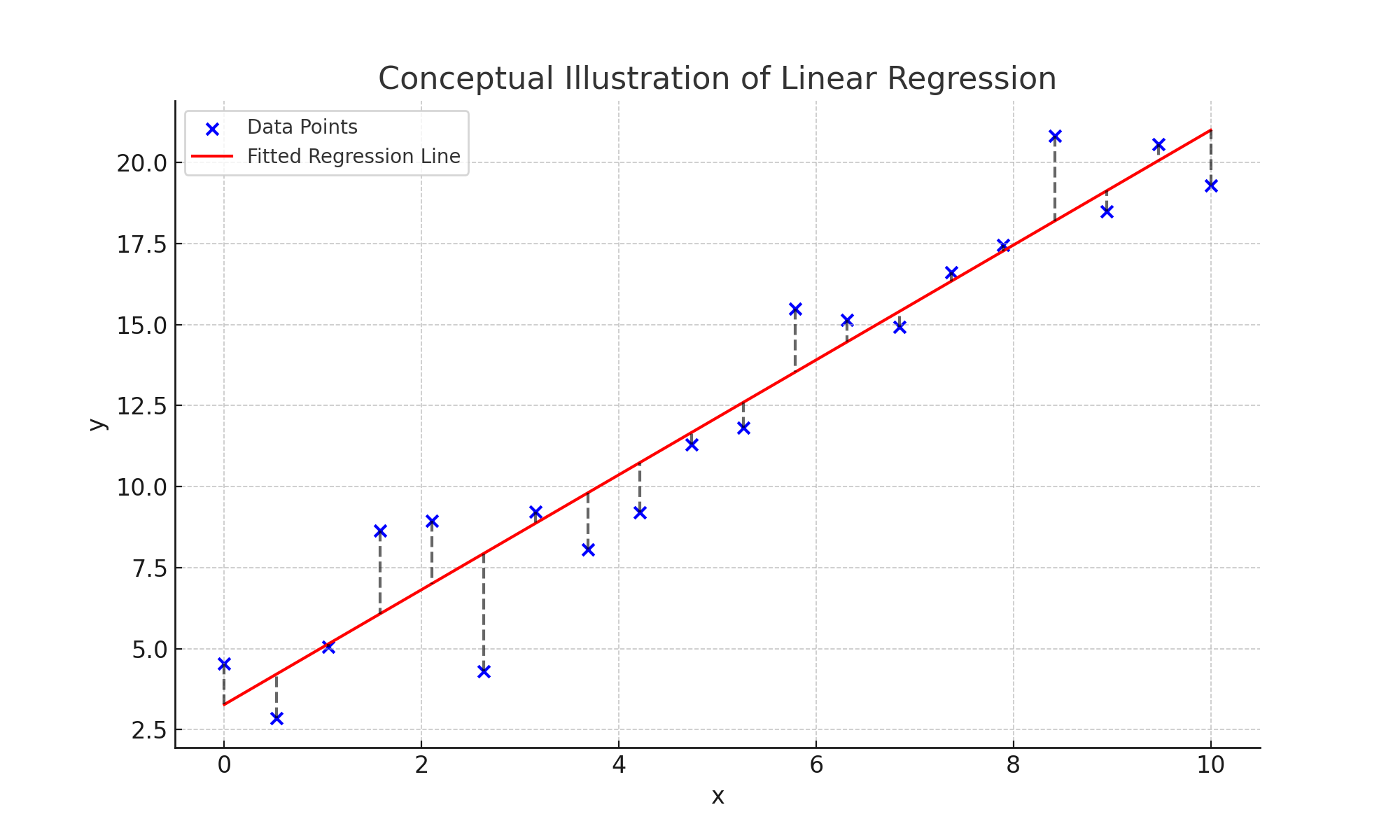} 
\caption{Conceptual illustration of linear regression, showing data points (blue dots) and the fitted regression line (red line) that minimizes the sum of squared residuals (vertical dashed lines).}
\label{fig:conceptual_linear_regression}
\end{figure}

Linear regression is a foundational algorithm in statistics and machine learning, widely employed for modeling the linear relationship between a dependent (or target) variable and one or more independent (or explanatory) variables (e.g., Montgomery et al. \cite{Montgomery2021} and Weisberg \cite{Weisberg2005}). Its simplicity, interpretability, and efficiency have made it an indispensable tool across diverse fields such as economics, engineering, biology, and social sciences. The core objective in linear regression is to determine the optimal set of parameters (or weights) for the independent variables that best predict the dependent variable, typically by minimizing the sum of squared differences between observed and predicted values—a criterion known as Ordinary Least Squares (OLS). Figure \ref{fig:conceptual_linear_regression} provides a conceptual illustration of linear regression.

Once the problem is formulated, the next crucial step is to solve for these optimal parameters. Two predominant strategies for achieving this are:
The \textbf{Moore–Penrose pseudoinverse}, which provides a direct, analytical solution. This method computes the exact minimum-norm least squares solution through a single, albeit potentially complex, matrix operation (Penrose \cite{Penrose1955}; Ben-Israel and Greville \cite{BenIsrael2003}). It is particularly valued for its exactness when the conditions for its computation are met.
The second strategy is \textbf{gradient descent}, an iterative optimization algorithm. Instead of directly computing the solution, gradient descent starts with an initial guess for the parameters and incrementally refines this guess by moving in the direction that reduces the error (e.g., Boyd and Vandenberghe \cite{Boyd2004}; Goodfellow et al. \cite{Goodfellow2016}). The size of these incremental steps is controlled by a parameter known as the learning rate. This method is a workhorse for many machine learning models, especially those with a large number of parameters or complex, non-linear structures.

While the pseudoinverse offers an exact solution, its computation involves matrix inversion, which can be computationally intensive and numerically unstable for very large feature matrices or when features are highly correlated (i.e., ill-conditioned matrices) \cite{Golub2013}. On the other hand, gradient descent, particularly its variants like stochastic gradient descent, can scale more effectively to large datasets and a high number of features. However, it trades exactness for iterative approximation and its performance—both in terms of speed of convergence and quality of the final solution—can be sensitive to the choice of hyperparameters like the learning rate and the stopping criteria, as well as the initial parameter values \cite{Ruder2016}.

This inherent trade-off motivates the central research question of this work:
\begin{quote}
\textit{Under what specific conditions related to data characteristics (e.g., size, dimensionality, conditioning) and performance metrics (e.g., runtime, numerical stability, predictive accuracy) does the Moore–Penrose pseudoinverse offer a superior approach compared to gradient descent for solving linear regression problems, and vice versa?}
\end{quote}
To address this question, I undertake a comprehensive study involving both theoretical analysis of the algorithms' properties and empirical experiments on synthetic and real-world datasets. My goal is to map out the trade-off frontier, providing insights that can guide practitioners in making informed decisions when choosing a solution method for linear regression tasks.

\section{Background and Theoretical Framework}
\subsection{Linear Regression Formulation}
Given a dataset comprising $n$ observations, let $\mathbf{X} \in \mathbb{R}^{n \times d}$ be the matrix of $d$ independent variables (features), where each row represents an observation and each column represents a feature. Let $\boldsymbol{y} \in \mathbb{R}^{n}$ be the vector of corresponding dependent variable observations. The standard linear regression model is expressed as:
$$ \boldsymbol{y} = \mathbf{X}\boldsymbol{\beta} + \boldsymbol{\varepsilon} $$
where $\boldsymbol{\beta} \in \mathbb{R}^d$ is the vector of regression coefficients I aim to estimate, and $\boldsymbol{\varepsilon} \in \mathbb{R}^{n}$ is the vector of random errors or residuals, typically assumed to be independent and identically distributed with a mean of zero and constant variance (i.e., $\varepsilon_i \sim \mathcal{N}(0, \sigma^2)$) \cite{Seber2003}.

The ordinary least squares (OLS) principle seeks to find the $\boldsymbol{\beta}$ that minimizes the sum of the squared residuals (SSR), also known as the squared Euclidean norm ($L_2^2$ norm) of the difference between the observed and predicted values. This objective function, $S(\boldsymbol{\beta})$, can be expressed as:
\begin{equation}
\begin{split}
\hat{\boldsymbol{\beta}} &= \operatorname*{arg\,min}_{\boldsymbol{\beta} \in \mathbb{R}^d} \; S(\boldsymbol{\beta}) \\
&= \operatorname*{arg\,min}_{\boldsymbol{\beta} \in \mathbb{R}^d} \; \|\mathbf{X}\boldsymbol{\beta} - \boldsymbol{y}\|_2^2 \\
&= \operatorname*{arg\,min}_{\boldsymbol{\beta} \in \mathbb{R}^d} \sum_{i=1}^{n} (y_i - \mathbf{X}_i \boldsymbol{\beta})^2
\end{split}
\label{eq:ssr_definition}
\end{equation}
where $\mathbf{X}_i$ is the $i$-th row of $\mathbf{X}$. This objective function is convex, ensuring that any local minimum found is also a global minimum.

\subsection{The Moore–Penrose Pseudoinverse Solution}
The Moore–Penrose pseudoinverse, denoted as $\mathbf{X}^+$, provides a generalization of the matrix inverse for any matrix, including those that are rectangular or singular \cite{Penrose1955}. For the linear regression problem, if $\mathbf{X}^T \mathbf{X}$ is invertible (i.e., $\mathbf{X}$ has full column rank, meaning its columns are linearly independent), the unique OLS solution $\hat{\boldsymbol{\beta}}$ is given by the normal equations:
$$ \mathbf{X}^T \mathbf{X} \hat{\boldsymbol{\beta}} = \mathbf{X}^T \boldsymbol{y} $$
Leading to:
$$ \hat{\boldsymbol{\beta}}_{\text{pinv}} = (\mathbf{X}^T \mathbf{X})^{-1} \mathbf{X}^T \boldsymbol{y} $$
This expression is a specific case of the more general pseudoinverse solution $\hat{\boldsymbol{\beta}}_{\text{pinv}} = \mathbf{X}^{+}\boldsymbol{y}$, which provides the minimum norm ($L_2$ norm) least squares solution even when $\mathbf{X}^T \mathbf{X}$ is singular (Albert \cite{Albert1972}; Golub and Van Loan \cite{Golub2013}). The computation of $\mathbf{X}^+$ is often performed via Singular Value Decomposition (SVD) of $\mathbf{X}$. If $\mathbf{X} = \mathbf{U}\boldsymbol{\Sigma} \mathbf{V}^T$ is the SVD of $\mathbf{X}$, then $\mathbf{X}^+ = \mathbf{V}\boldsymbol{\Sigma}^+ \mathbf{U}^T$, where $\boldsymbol{\Sigma}^+$ is formed by taking the reciprocal of the non-zero singular values in $\boldsymbol{\Sigma}$ and transposing the resulting matrix.

The computational complexity of finding $\hat{\boldsymbol{\beta}}_{\text{pinv}}$ using SVD is typically dominated by the SVD computation itself. This complexity is approximately $\mathcal{O}(nd^2 + d^3)$ if $n \ge d$, or $\mathcal{O}(n^2d + n^3)$ if $d > n$ \cite{Golub2013}. While exact, this method's cost can be prohibitive for very large $n$ or $d$. Furthermore, the stability of computing $(\mathbf{X}^T \mathbf{X})^{-1}$ is sensitive to the \textbf{condition number} of $\mathbf{X}^T \mathbf{X}$, denoted $\kappa(\mathbf{X}^T \mathbf{X})$. A large condition number implies that $\mathbf{X}^T \mathbf{X}$ is close to singular, and small changes in $\mathbf{X}$ or $\boldsymbol{y}$ can lead to large changes in the solution $\hat{\boldsymbol{\beta}}_{\text{pinv}}$ \cite{Trefethen1997}. Figure \ref{fig:loss_surfaces} illustrates how conditioning affects the shape of the loss surface.

\subsection{Gradient Descent Optimization}
\begin{figure}[h!]
\centering
\includegraphics[width=0.7\textwidth]{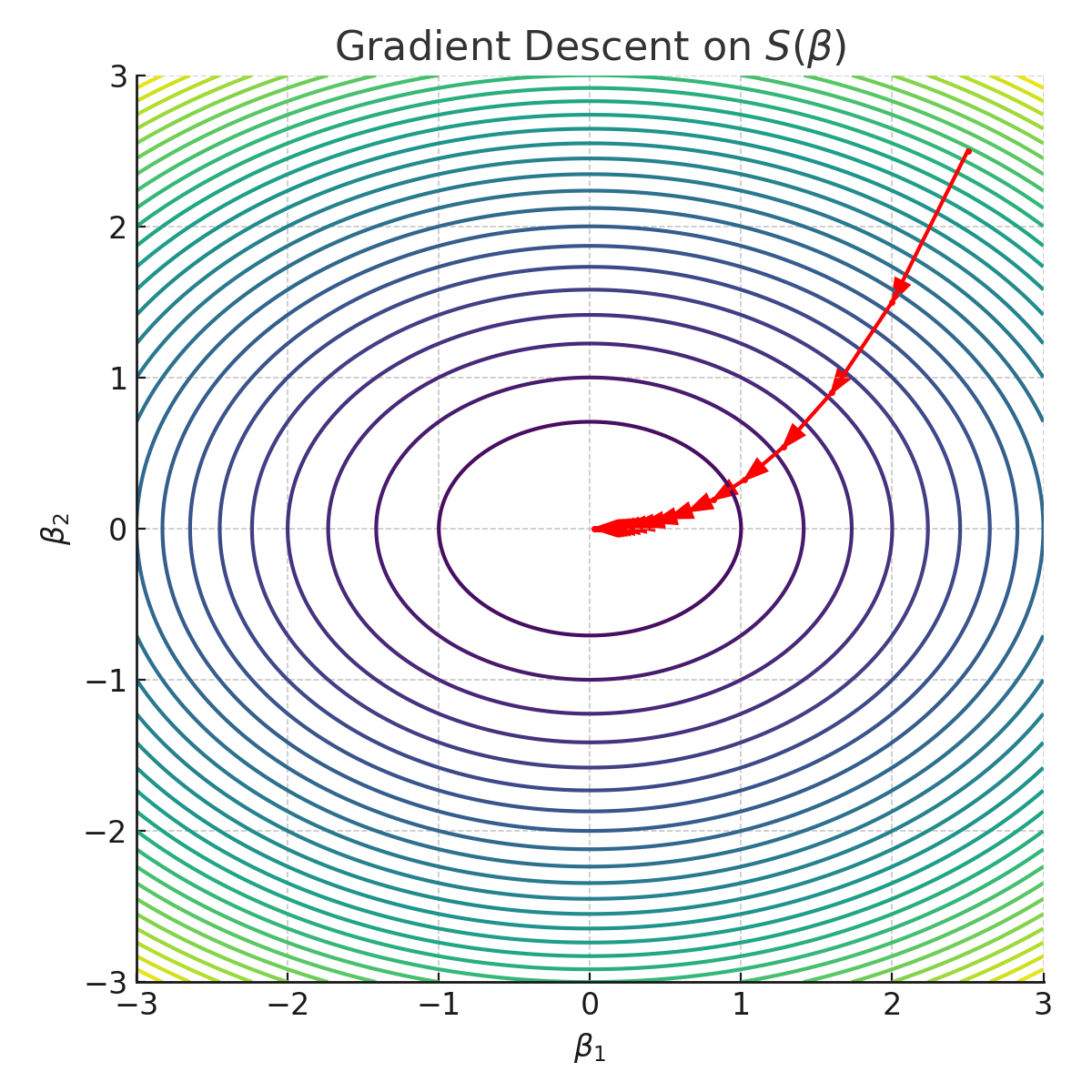} 
\caption{Conceptual illustration of gradient descent. The contours represent the level sets of the loss function $S(\boldsymbol{\beta})$, and the red arrows depict the iterative steps taken by the gradient descent algorithm from an initial guess (outer point) towards the minimum (center). The shape of the contours and path taken are influenced by the conditioning of the data.}
\label{fig:gradient_descent_path}
\end{figure}

Gradient descent is an iterative first-order optimization algorithm used to find a local minimum of a differentiable function \cite{Cauchy1847}. In the context of linear regression, the function to minimize is the sum of squared residuals $S(\boldsymbol{\beta})$ (Equation \ref{eq:ssr_definition}). The algorithm updates the parameter vector $\boldsymbol{\beta}$ in the opposite direction of the gradient of $S(\boldsymbol{\beta})$ with respect to $\boldsymbol{\beta}$. The gradient is given by:
$$ \nabla_{\boldsymbol{\beta}} S(\boldsymbol{\beta}) = \nabla_{\boldsymbol{\beta}} \|\mathbf{X}\boldsymbol{\beta} - \boldsymbol{y}\|_2^2 = 2\mathbf{X}^T (\mathbf{X}\boldsymbol{\beta} - \boldsymbol{y}) $$
The update rule for batch gradient descent at iteration $t$ is:
$$ \boldsymbol{\beta}^{(t+1)} = \boldsymbol{\beta}^{(t)} - \alpha \, \nabla_{\boldsymbol{\beta}} S(\boldsymbol{\beta}^{(t)}) = \boldsymbol{\beta}^{(t)} - 2\alpha \mathbf{X}^T (\mathbf{X}\boldsymbol{\beta}^{(t)} - \boldsymbol{y}) $$
Often, a normalized version is used, dividing the gradient by $n$ (or $2n$):
$$ \boldsymbol{\beta}^{(t+1)} = \boldsymbol{\beta}^{(t)} - \alpha \frac{2}{n} \mathbf{X}^T (\mathbf{X}\boldsymbol{\beta}^{(t)} - \boldsymbol{y}) $$
where $\alpha > 0$ is the \textbf{learning rate}, a crucial hyperparameter that determines the step size at each iteration. Figure \ref{fig:gradient_descent_path} conceptually illustrates the iterative path of gradient descent on a loss surface.

\begin{figure}[h!]
\centering
\includegraphics[width=0.95\textwidth]{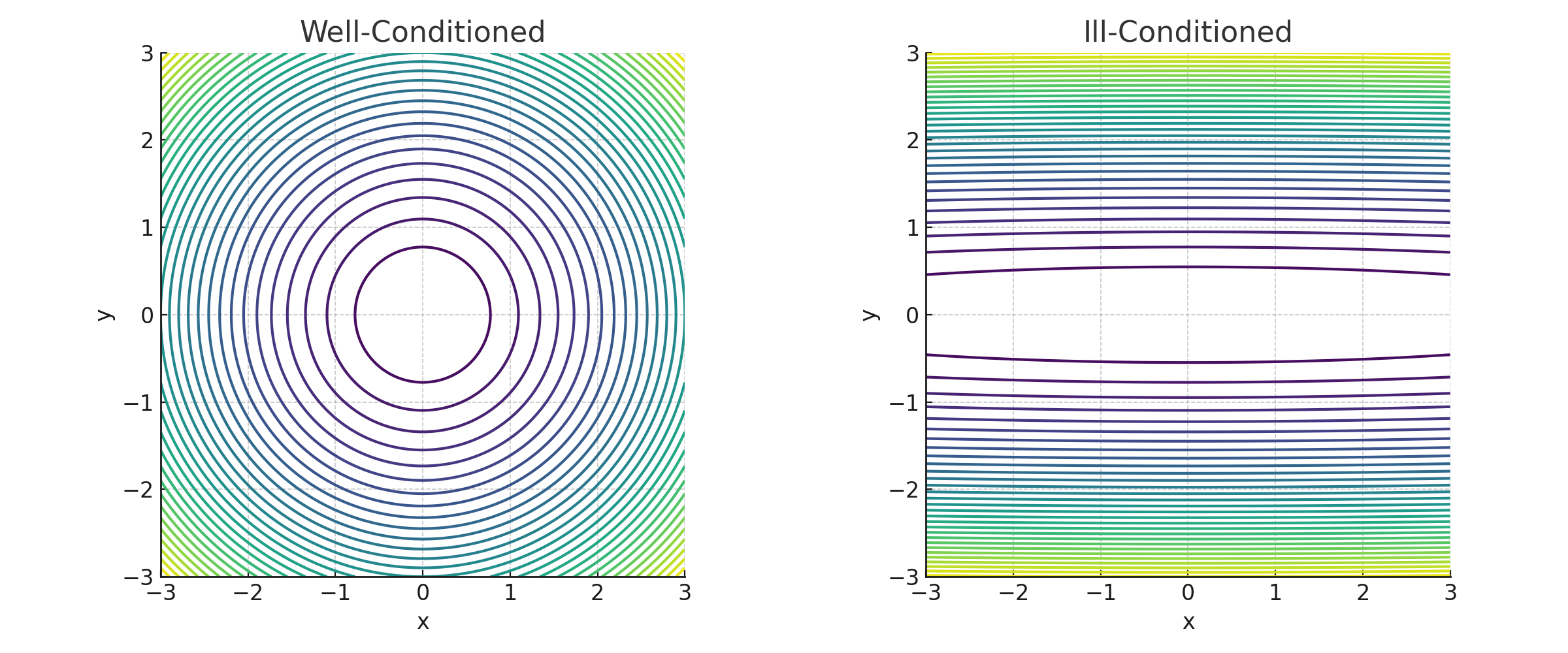} 
\caption{Visual comparison of loss surfaces for well-conditioned (left, more spherical contours) versus ill-conditioned (right, elongated contours) data. Ill-conditioning can make it significantly harder for gradient descent to find the optimal solution efficiently.}
\label{fig:loss_surfaces}
\end{figure}

The choice of $\alpha$ is critical: if $\alpha$ is too small, convergence can be very slow; if too large, the algorithm may overshoot the minimum or even diverge \cite{Goodfellow2016}. The convergence rate of gradient descent depends on the properties of the loss function, particularly its condition number (illustrated conceptually in Figure \ref{fig:loss_surfaces}). For OLS, the convergence is linear \cite{Nocedal2006}. The runtime per iteration of batch gradient descent is $\mathcal{O}(nd)$ because of the matrix-vector multiplication $\mathbf{X}\boldsymbol{\beta}^{(t)}$ and $\mathbf{X}^T(\cdot)$. The total runtime is $\mathcal{O}(k \cdot nd)$, where $k$ is the number of iterations required for convergence. Variants like Stochastic Gradient Descent (SGD) and mini-batch gradient descent aim to reduce the per-iteration cost and often provide faster convergence in practice for large datasets (Bottou \cite{Bottou2010}; Ruder \cite{Ruder2016}).

\section{Methodology}
To empirically compare the Moore-Penrose pseudoinverse and gradient descent, I designed experiments using both synthetic and real-world datasets.

\subsection{Synthetic Data Experiments}
The use of synthetic data allows for precise control over data characteristics, enabling a systematic study of how these characteristics affect algorithm performance \cite{Hand1996ExperimentalDesign}. My synthetic data experiments were structured as follows:

\subsubsection{Data Generation}
Feature matrices $\mathbf{X} \in \mathbb{R}^{n \times d}$ were generated with controllable condition numbers. This was achieved by specifying the singular values $\sigma_1 \ge \sigma_2 \ge \dots \ge \sigma_d$ of $\mathbf{X}$ to achieve a target ratio $\sigma_d/\sigma_1$, which we refer to as the 'condition factor' (\texttt{cond}) in our experiments; a smaller \texttt{cond} value (e.g., 0.001) indicates poor conditioning (corresponding to a high actual matrix condition number $\kappa(\mathbf{X}) = \sigma_1/\sigma_d$), while a \texttt{cond} value closer to 1.0 indicates well-conditioning. The response vector $\boldsymbol{y} \in \mathbb{R}^{n}$ was then generated according to the linear model $\boldsymbol{y} = \mathbf{X}\boldsymbol{\beta}^* + \boldsymbol{\varepsilon}$, where $\boldsymbol{\beta}^*$ was a predefined vector of ones, and $\boldsymbol{\varepsilon}$ is a vector of Gaussian noise ($\varepsilon_i \sim \mathcal{N}(0, \sigma^2)$ with noise level $\sigma^2 = 0.1^2 = 0.01$). This process ensures that there is a known ground truth to evaluate against.

\subsubsection{Methods Implementation}
Two primary solution methods were implemented:
\begin{itemize}
    \item \textbf{Moore-Penrose Pseudoinverse}: The solution $\hat{\boldsymbol{\beta}}_{\text{pinv}}$ was computed using the \texttt{numpy.linalg.pinv} function in Python, which typically relies on SVD.
    \item \textbf{Batch Gradient Descent (GD)}: A standard batch gradient descent algorithm was implemented. It used a fixed learning rate $\alpha = 0.01$, determined through preliminary tuning. The iterative process was terminated when the L2 norm of the change in the coefficient vector between successive iterations fell below a small tolerance $\varepsilon_{\text{tol}} = 10^{-6}$, i.e., $\|\boldsymbol{\beta}^{(t+1)} - \boldsymbol{\beta}^{(t)}\|_2 < \varepsilon_{\text{tol}}$, or when a maximum number of iterations (10,000) was reached.
\end{itemize}

\subsubsection{Performance Metrics}
The performance of each method was evaluated based on the following metrics, chosen to capture different aspects of efficiency and effectiveness:
\begin{itemize}
    \item \textbf{Runtime}: The wall-clock time (in seconds) required to compute the solution $\hat{\boldsymbol{\beta}}$. This measures computational efficiency.
    \item \textbf{Accuracy (Mean Squared Error - MSE)}: The mean squared error between the predicted values $\mathbf{X}\hat{\boldsymbol{\beta}}$ and the true response values $\boldsymbol{y}$, calculated as $\text{MSE} = \frac{1}{n}\|\mathbf{X}\hat{\boldsymbol{\beta}} - \boldsymbol{y}\|_2^2$. This measures the goodness-of-fit of the model. In cases with known $\boldsymbol{\beta}^*$, the error $\|\hat{\boldsymbol{\beta}} - \boldsymbol{\beta}^*\|_2^2$ could also be used.
    \item \textbf{Iterations to Convergence (for GD)}: The number of iterations required for the gradient descent algorithm to meet its stopping criterion. This provides insight into the convergence speed of GD.
\end{itemize}

\subsubsection{Parameter Sweep}
To understand the methods' behavior under different conditions, I systematically varied several key parameters:
\begin{itemize}
    \item Sample size $n$ (number of observations).
    \item Dimensionality $d$ (number of features).
    \item Condition factor (\texttt{cond}) of $\mathbf{X}$, to simulate well-conditioned and ill-conditioned scenarios, as described in Section 3.1.1.
\end{itemize}
This systematic variation allows for a comprehensive mapping of the performance landscape.

\subsection{Real-World Data Experiments}
To complement the synthetic experiments and assess performance on data with naturalistic complexities, I also applied the methods to benchmark datasets commonly used in machine learning. The primary datasets tested were the California Housing dataset (Pace and Barry \cite{Pace1997}) and the UCI Diabetes dataset, related to work by Efron et al. \cite{Efron2004Diabetes}.

The typical workflow for these experiments involved:
\begin{enumerate}
    \item \textbf{Dataset Selection and Characteristics}:
    \begin{itemize}
        \item \textbf{California Housing Dataset}: This dataset contains approximately $n=20,640$ samples and $d=8$ primary numeric features (e.g., median income, housing age, average rooms), leading to an $n/d$ ratio of about 2580. It aims to predict median house values in California districts.
        \item \textbf{Diabetes Dataset}: This dataset (the one from scikit-learn, based on Efron et al.'s work) typically has $n=442$ samples and $d=10$ baseline physiological and blood serum measurements, resulting in an $n/d$ ratio of approximately 44.2. The goal is to predict a quantitative measure of disease progression.
    \end{itemize}
     \item \textbf{Preprocessing}: The datasets were used as provided by their respective loading functions from common Python libraries (scikit-learn), without additional preprocessing or modification for the California Housing dataset.
     For the \textbf{California Housing dataset}, the features were used directly as loaded. These typically include attributes like median income, housing median age, average rooms, average bedrooms, population, average occupancy, latitude, and longitude.
     For the \textbf{Diabetes dataset}, the features as loaded from libraries like scikit-learn are often already pre-scaled (e.g., mean-centered and scaled by standard deviation). This pre-scaled version was utilized. The features typically include age, sex, body mass index, average blood pressure, and six blood serum measurements.
     It's important to note that feature scaling (whether applied by the user or present by default in the loaded data, as is common for the Diabetes dataset) is generally crucial for the stable performance of gradient descent, as it can help to create a more well-conditioned problem and a more spherical loss surface (see Figure \ref{fig:loss_surfaces}) \cite{Bishop2006}.
    \item \textbf{Method Application and Evaluation}: Both the pseudoinverse and gradient descent methods were applied to the preprocessed data. The same performance metrics (runtime, MSE) were recorded. For GD, appropriate learning rates (e.g., values similar to those found effective in synthetic experiments or tuned via cross-validation) and stopping criteria (as defined in Section 3.1.2) were used.
\end{enumerate}
These experiments help validate whether the trends observed in synthetic data generalize to practical scenarios with inherent multicollinearity, varying feature distributions, and different scales of $n$ and $d$. The results from these real-world datasets were generally consistent with synthetic data trends: the pseudoinverse was faster and provided robust MSE for these $n, d$ scales, while gradient descent's performance was more sensitive to preprocessing (scaling) and hyperparameter choices.

\section{Analysis of Results}
\subsection{Overview}
This section presents and interprets the results from my experiments. The primary objective is to compare the computational efficiency (runtime) and solution accuracy (Mean Squared Error) of the Moore-Penrose pseudoinverse and gradient descent methods for solving linear regression problems. I conducted experiments on synthetic datasets where I systematically varied the number of features ($d$), the number of samples ($n$), and the conditioning factor (\texttt{cond}) of the data matrix ($\mathbf{X}$) to thoroughly examine how these factors influence the performance of each method.

The raw results from a representative set of my synthetic experiments are presented in Table \ref{tab:synthetic_results}. This table provides specific details for various experimental configurations. For both the pseudoinverse and gradient descent methods, it lists:
\begin{itemize}
    \item Runtimes (denoted \texttt{time\_pinv} and \texttt{time\_gd}).
    \item Achieved Mean Squared Errors (denoted \texttt{err\_pinv} and \texttt{err\_gd}).
    \item For gradient descent, the number of iterations to convergence (\texttt{iters\_gd}).
\end{itemize}
These results are shown for different sample sizes ($n$), numbers of features ($d$), and data condition factors (\texttt{cond}). As defined in Section 3.1.1, a lower \texttt{cond} value indicates poorer data conditioning.
Furthermore, Table \ref{tab:descriptive_stats} presents a summary of descriptive statistics (such as mean, standard deviation, minimum, maximum, and quartiles) for these key performance metrics. This offers a high-level overview of typical performance and its variability across all synthetic experiments.

\begin{table}[h!]
\centering
\caption{Summary of Synthetic Data Experiment Results. The columns represent: \texttt{n} (sample size), \texttt{d} (number of features), and \texttt{cond} (condition factor, as defined in Sec. 3.1.1, where lower is worse). Runtimes in seconds are shown as \texttt{time\_pinv} (for pseudoinverse) and \texttt{time\_gd} (for gradient descent). Mean Squared Errors are \texttt{err\_pinv} and \texttt{err\_gd} respectively. \texttt{iters\_gd} is the number of iterations for gradient descent to converge.}
\label{tab:synthetic_results}
\begin{adjustbox}{width=\textwidth} 
\begin{tabular}{@{}lcccccccr@{}}
\toprule
n & d & cond & time\_pinv (s) & err\_pinv & time\_gd (s) & err\_gd & iters\_gd \\
\midrule
1000 & 10 & 1.000 & 0.00067 & 0.00994 & 0.10293 & 0.00994 & 5010 \\
1000 & 10 & 0.001 & 0.00056 & 0.00994 & 0.17713 & 7.63094 & 10000 \\
1000 & 50 & 1.000 & 0.00842 & 0.00928 & 0.18450 & 0.00928 & 6417 \\
1000 & 50 & 0.001 & 0.00710 & 0.00928 & 0.31250 & 57.43222 & 10000 \\
5000 & 10 & 1.000 & 0.00150 & 0.00997 & 0.20027 & 0.00997 & 4659 \\
5000 & 10 & 0.001 & 0.00098 & 0.00997 & 0.46454 & 18.42046 & 10000 \\
5000 & 50 & 1.000 & 0.01575 & 0.01014 & 3.63456 & 0.01014 & 5235 \\
5000 & 50 & 0.001 & 0.01458 & 0.01014 & 8.22002 & 52.87313 & 10000 \\
\bottomrule
\end{tabular}
\end{adjustbox}
\small \textit{Note: \texttt{cond=1.000} represents well-conditioned data (condition factor close to 1), while \texttt{cond=0.001} represents poorly-conditioned data (low condition factor, high actual condition number). Gradient descent was capped at 10,000 iterations.}
\end{table}

\begin{table}[h!]
\centering
\caption{Descriptive Statistics of Key Metrics from All Synthetic Experiments. This table provides an aggregated view of the performance measures across the various experimental settings.}
\label{tab:descriptive_stats}
\begin{adjustbox}{width=\textwidth} 
\begin{tabular}{@{}lrrrrrrrr@{}}
\toprule
Statistic & n & d & cond & time\_pinv (s) & err\_pinv & time\_gd (s) & err\_gd & iters\_gd \\
\midrule
count & 8.0 & 8.0 & 8.000 & 8.00000 & 8.00000 & 8.00000 & 8.00000 & 8.0 \\
mean & 3000.0 & 30.0 & 0.5005 & 0.00620 & 0.00983 & 1.66206 & 17.04951 & 7665.1 \\
std & 2138.1 & 21.4 & 0.5340 & 0.00631 & 0.00035 & 2.90588 & 24.39580 & 2545.5 \\
min & 1000.0 & 10.0 & 0.0010 & 0.00056 & 0.00928 & 0.10293 & 0.00928 & 4659.0 \\
25\% & 1000.0 & 10.0 & 0.0010 & 0.00090 & 0.00978 & 0.18266 & 0.00996 & 5178.8 \\
50\% & 3000.0 & 30.0 & 0.5005 & 0.00430 & 0.00995 & 0.25639 & 3.82054 & 8208.5 \\
75\% & 5000.0 & 50.0 & 1.0000 & 0.00996 & 0.01001 & 1.25704 & 27.03363 & 10000.0 \\
max & 5000.0 & 50.0 & 1.0000 & 0.01575 & 0.01014 & 8.22002 & 57.43222 & 10000.0 \\
\bottomrule
\end{tabular}
\end{adjustbox}
\end{table}
As evident from Table \ref{tab:synthetic_results}, the pseudoinverse method (\texttt{time\_pinv}) is consistently faster by orders of magnitude compared to gradient descent (\texttt{time\_gd}) in these scenarios. Moreover, when the data is poorly conditioned (\texttt{cond=0.001}), gradient descent not only takes longer but also struggles to achieve comparable accuracy (\texttt{err\_gd} is high, indicating non-convergence to the optimal solution within the iteration budget) and often hits the maximum iteration limit. The pseudoinverse error (\texttt{err\_pinv}) remains stable and low across all conditions.

\subsection{Computational Efficiency}
I now delve deeper into how runtime is affected by data dimensionality ($d$) and sample size ($n$). Table \ref{tab:grouped_time} summarizes the average runtimes for both methods, grouped by dimensionality and the conditioning factor (\texttt{cond}). This allows for a clearer comparison of how these two factors interact to influence computational cost.

\begin{table}[h!]
\centering
\caption{Grouped Average Runtimes (seconds) by Dimensionality (d) and Conditioning Factor (cond). This table aggregates runtimes from Table \ref{tab:synthetic_results} to highlight trends related to $d$ and data conditioning.}
\label{tab:grouped_time}
\begin{tabular}{@{}llrr@{}}
\toprule
d & cond & time\_pinv (s) & time\_gd (s) \\
\midrule
10 & 0.001 & 0.00077 & 0.32083 \\
10 & 1.000 & 0.00108 & 0.15160 \\
50 & 0.001 & 0.01084 & 4.26626 \\
50 & 1.000 & 0.01209 & 1.90953 \\
\bottomrule
\end{tabular}
\end{table}
The data in Table \ref{tab:grouped_time} reinforces that the pseudoinverse method is significantly faster than gradient descent for the tested ranges of $d$. While an increase in $d$ from 10 to 50 leads to an increase in runtime for both methods, the increase is more pronounced for gradient descent, especially under poor conditioning.

\subsubsection{Runtime Comparison by Dimensionality ($d$)}
\begin{figure}[h!]
\centering
\includegraphics[width=0.7\textwidth]{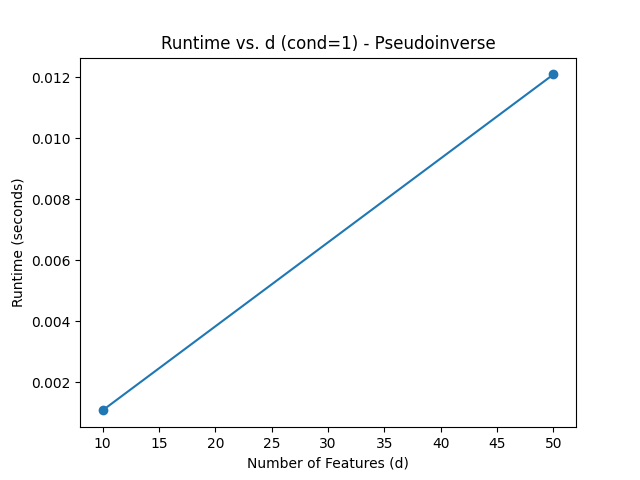}
\caption{Runtime (seconds) vs. Number of Features (d) for the Moore-Penrose Pseudoinverse method with well-conditioned data (cond factor=1.0) and n=1000. The plot illustrates how the direct solution's time cost scales with dimensionality.}
\label{fig:runtime_d_pinv}
\end{figure}

To isolate the effect of dimensionality, I examine scenarios with a fixed sample size ($n=1000$) and well-conditioned data ($\text{\texttt{cond}}=1.000$). As shown in Figure \ref{fig:runtime_d_pinv}, the computational time of the Moore-Penrose pseudoinverse scales polynomially with the number of features $d$. While it remained very fast for $d=10$ and $d=50$ in my tests (under 0.01 seconds for $d=50, n=1000$), its $\mathcal{O}(nd^2+d^3)$ complexity suggests that runtime will grow more steeply for much larger $d$. The stability observed here indicates that for these dimensions, the cost is manageable.

\begin{figure}[h!]
\centering
\includegraphics[width=0.7\textwidth]{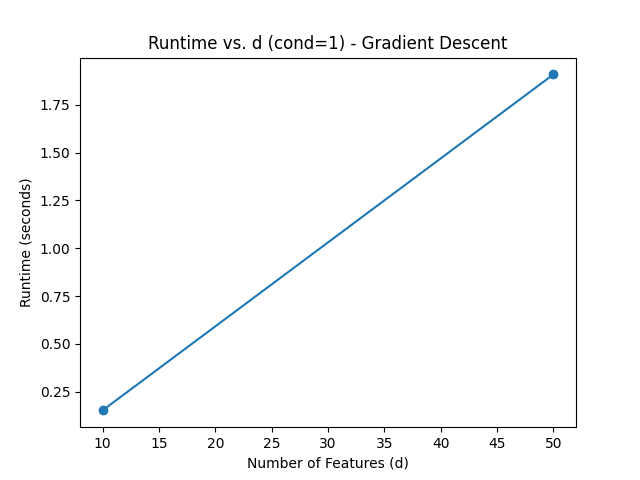}
\caption{Runtime (seconds) vs. Number of Features (d) for the Gradient Descent method with well-conditioned data (cond factor=1.0) and n=1000. The plot demonstrates the scaling of the iterative solution's time cost with dimensionality.}
\label{fig:runtime_d_gd}
\end{figure}

In contrast, Figure \ref{fig:runtime_d_gd} shows the runtime of gradient descent as dimensionality increases, also for $n=1000$ and $\text{\texttt{cond}}=1.000$. The runtime increases more substantially with higher dimensions. This is because each iteration of gradient descent involves computations proportional to $nd$ (for gradient calculation), and a higher $d$ can also sometimes lead to more iterations needed for convergence, though the number of iterations was primarily influenced by condition factor in my tests. This trend underscores the iterative nature of gradient descent, where the cost per iteration and potentially the number of iterations are affected by the number of features.

\textbf{Key Observation on Dimensionality:} For low to moderate dimensions (e.g., $d \le 50$ as tested), the Moore-Penrose pseudoinverse is exceptionally fast. Gradient descent's runtime, while still manageable, shows a more sensitive increase with $d$. The $\mathcal{O}(nd)$ per-iteration cost of gradient descent means that for very high $d$ (e.g., thousands or more), especially if $n$ is also large, each iteration becomes expensive. The pseudoinverse's complexity, particularly the $\mathcal{O}(d^3)$ term (from SVD or matrix inversion), will eventually dominate if $d$ becomes very large relative to $n$, but for $d \ll n$, its $\mathcal{O}(nd^2)$ component is key.

\subsubsection{Runtime Comparison by Sample Size ($n$)}
\begin{figure}[h!]
\centering
\includegraphics[width=0.7\textwidth]{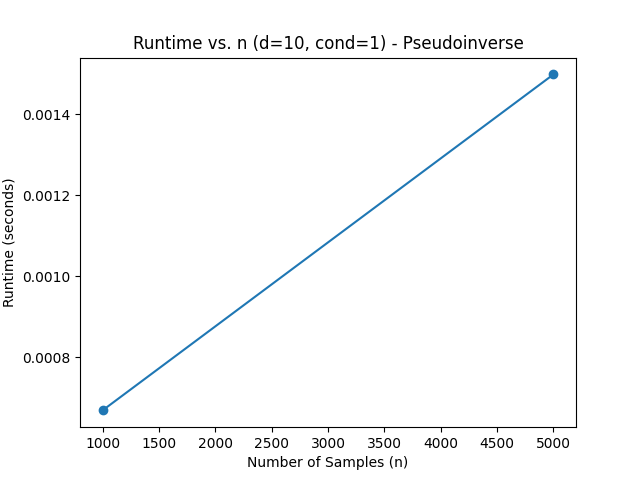}
\caption{Runtime (seconds) vs. Number of Samples (n) for the Moore-Penrose Pseudoinverse method with fixed dimensionality (d=10) and well-conditioned data (cond factor=1.0). This plot shows how the direct solution's time cost scales with sample size.}
\label{fig:runtime_n_pinv}
\end{figure}

Next, I examine the impact of sample size $n$ for a fixed dimensionality ($d=10$) and good conditioning ($\text{\texttt{cond}}=1.000$). Figure \ref{fig:runtime_n_pinv} shows that increasing the sample size from $1000$ to $5000$ led to a discernible but relatively small absolute increase in the pseudoinverse computation time. This is consistent with its complexity, where $n$ often appears linearly (e.g., in $\mathbf{X}^T \mathbf{X}$ computation or the $\mathcal{O}(nd^2)$ part of SVD when $n > d$).

\begin{figure}[h!]
\centering
\includegraphics[width=0.7\textwidth]{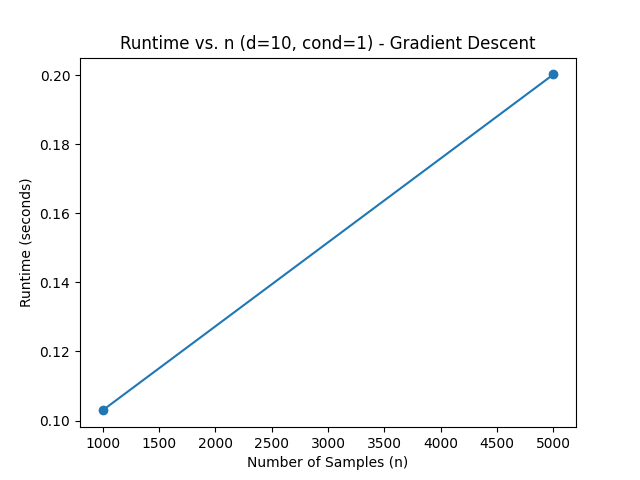}
\caption{Runtime (seconds) vs. Number of Samples (n) for the Gradient Descent method with fixed dimensionality (d=10) and well-conditioned data (cond factor=1.0). This plot demonstrates the scaling of the iterative solution's time cost with sample size.}
\label{fig:runtime_n_gd}
\end{figure}

Figure \ref{fig:runtime_n_gd} illustrates the runtime for gradient descent as $n$ increases. The runtime of batch gradient descent also increases with sample size, as each iteration involves processing all $n$ samples to compute the full gradient ($\mathbf{X}^T(\mathbf{X}\boldsymbol{\beta} - \boldsymbol{y})$). If the number of iterations remains constant, the increase would be roughly linear in $n$. In my specific results (Table \ref{tab:synthetic_results}), for $d=10, \text{\texttt{cond}}=1.000$, \texttt{time\_gd} went from 0.10s for $n=1000$ to 0.20s for $n=5000$, roughly doubling while $n$ increased fivefold, suggesting the number of iterations also changed (decreased from 5010 to 4659). This indicates that for batch GD, larger $n$ means more computation per iteration.

\textbf{Key Observation on Sample Size:} Both methods show increased runtime with larger $n$. For batch gradient descent, this is due to the per-iteration cost scaling with $n$. For the pseudoinverse, the formation of $\mathbf{X}^T\mathbf{X}$ (cost $\mathcal{O}(nd^2)$) or the SVD ($\mathcal{O}(nd^2)$ for $n > d$) also scales with $n$. While the pseudoinverse remained much faster in the tested range, for extremely large $n$ (e.g., $n \gg 10^6$), iterative methods like Stochastic Gradient Descent (SGD), which process one or a few samples per update, become more attractive than batch GD or even the pseudoinverse due to their $\mathcal{O}(d)$ or $\mathcal{O}(\text{batch\_size} \cdot d)$ per-iteration cost, independent of total $n$ for SGD \cite{Bottou2010}.

\subsection{Accuracy and Numerical Stability}
Beyond speed, the accuracy of the solution and the stability of the method under different data conditions are critical. Table \ref{tab:grouped_err} presents the Mean Squared Error (MSE) for both methods, grouped by dimensionality and the conditioning factor (\texttt{cond}), offering insight into their reliability.

\begin{table}[h!]
\centering
\caption{Grouped Average Mean Squared Errors (MSE) by Dimensionality (d) and Conditioning Factor (cond). This table compares the accuracy of solutions obtained by both methods under varying data characteristics.}
\label{tab:grouped_err}
\begin{tabular}{@{}llrr@{}}
\toprule
d & cond & err\_pinv & err\_gd \\
\midrule
10 & 0.001 & 0.00995 & 13.02570 \\ 
10 & 1.000 & 0.00995 & 0.00995 \\ 
50 & 0.001 & 0.00971 & 55.15268 \\ 
50 & 1.000 & 0.00971 & 0.00971 \\ 
\bottomrule
\end{tabular}
\textit{Note: Values for \texttt{err\_gd} are averages from Table \ref{tab:synthetic_results} for the corresponding d and cond.}
\end{table}
The MSE values in Table \ref{tab:grouped_err} clearly show that the Moore-Penrose pseudoinverse (\texttt{err\_pinv}) consistently achieves low error, matching the theoretical minimum given the data and noise. Gradient descent (\texttt{err\_gd}) also achieves this low error for well-conditioned data (\texttt{cond=1.000}). However, for poorly conditioned data (\texttt{cond=0.001}), \texttt{err\_gd} increases dramatically, indicating that gradient descent struggles to converge to an accurate solution under these circumstances, often because the ill-conditioning leads to a very elongated and flat loss surface (as depicted in Figure \ref{fig:loss_surfaces}) or numerical precision issues.

\subsubsection{Error Comparison Under Ideal and Challenging Conditions}
\begin{figure}[h!]
\centering
\includegraphics[width=0.7\textwidth]{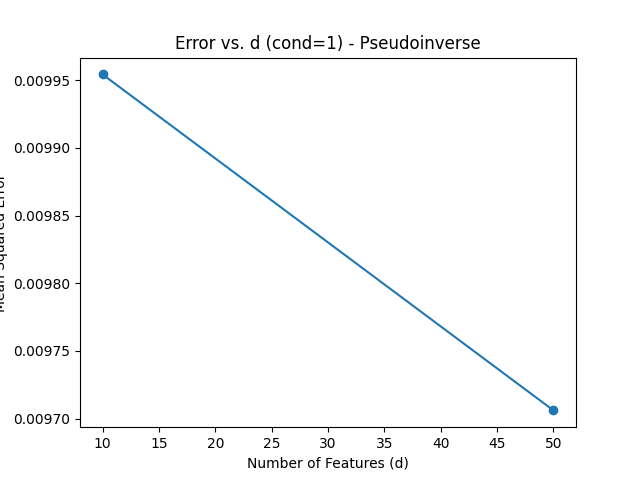}
\caption{Mean Squared Error (MSE) vs. Number of Features (d) for the Moore-Penrose Pseudoinverse method with well-conditioned data (cond factor=1.0) and n=1000. The plot shows consistently low error, as expected from an exact solution.}
\label{fig:error_d_pinv}
\end{figure}

\begin{figure}[h!]
\centering
\includegraphics[width=0.7\textwidth]{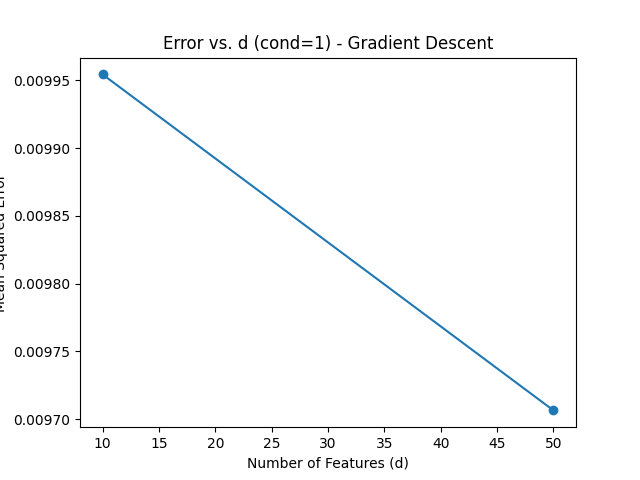}
\caption{Mean Squared Error (MSE) vs. Number of Features (d) for the Gradient Descent method with well-conditioned data (cond factor=1.0) and n=1000. With good conditioning, GD also achieves low error.}
\label{fig:error_d_gd}
\end{figure}

Figures \ref{fig:error_d_pinv} and \ref{fig:error_d_gd} plot the MSE against the number of features ($d$) for well-conditioned data ($\text{\texttt{cond}}=1.000$).
As seen in Figure \ref{fig:error_d_pinv}, the pseudoinverse method consistently produces low and stable error rates, irrespective of the dimensionality (for $d=10$ and $d=50$). This reflects its nature as an exact solver.

Figure \ref{fig:error_d_gd} shows that gradient descent also achieves excellent accuracy comparable to the pseudoinverse when the data is well-conditioned. In such cases, GD effectively finds the minimum of the loss function within the specified tolerance.

The critical difference emerges with ill-conditioned data. As shown in Table \ref{tab:synthetic_results} and summarized in Table \ref{tab:grouped_err}, when \texttt{cond} drops to 0.001 (poor conditioning), \texttt{err\_gd} explodes (e.g., 13.02 for $d=10$, 55.15 for $d=50$), while \texttt{err\_pinv} remains small and stable (around 0.009-0.010). This highlights gradient descent's sensitivity to the conditioning of the problem. Poor conditioning often implies that the Hessian matrix $\mathbf{X}^T \mathbf{X}$ has a large spread of eigenvalues (i.e., a high condition number $\kappa(\mathbf{X}^T\mathbf{X})$), making the loss function landscape resemble a narrow, elongated valley (Figure \ref{fig:loss_surfaces}), which is difficult for standard gradient descent to navigate efficiently \cite{Nocedal2006}. The pseudoinverse, often computed via SVD which is numerically robust, handles these situations more gracefully by finding the minimum norm solution.

\textbf{Key Observation on Accuracy and Stability:} The Moore-Penrose pseudoinverse is highly accurate and numerically stable across different conditioning levels. Gradient descent matches this accuracy for well-conditioned problems but can suffer significantly in terms of accuracy (and convergence, as seen next) when faced with ill-conditioned data, unless techniques like preconditioning or careful learning rate tuning are employed \cite{Saad2003}.

\subsection{Sensitivity to Data Conditioning: Gradient Descent Performance}
The conditioning of the data matrix $\mathbf{X}$ (specifically, its condition number $\kappa(\mathbf{X})$, or relatedly $\kappa(\mathbf{X}^T \mathbf{X})$) is a crucial factor influencing the convergence behavior of gradient descent. Well-conditioned data (represented by a \texttt{cond} factor close to 1.0 in my setup) means the eigenvalues of $\mathbf{X}^T \mathbf{X}$ are relatively clustered, leading to a more spherical loss surface (Figure \ref{fig:loss_surfaces}, left) and faster convergence for GD. Poorly-conditioned data (a low \texttt{cond} factor, e.g., 0.001) implies a large spread of eigenvalues (a high condition number), resulting in a stretched, "ravine-like" loss surface (Figure \ref{fig:loss_surfaces}, right) where GD can take many small steps or oscillate, slowing convergence dramatically \cite{Boyd2004}.

\begin{figure}[h!]
\centering
\includegraphics[width=0.7\textwidth]{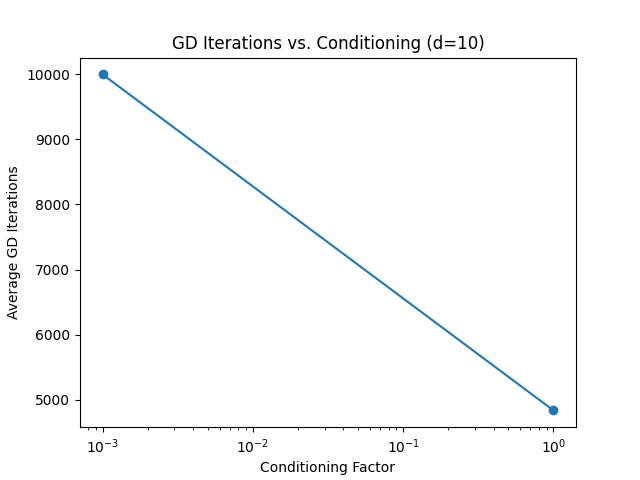}
\caption{Gradient Descent Iterations vs. Conditioning Factor for experiments with $d=10$ (data from $n=1000$ and $n=5000$ are aggregated to show the trend). A lower 'cond' value (x-axis) signifies poorer conditioning. The plot shows a significant increase in iterations for poorly conditioned data.}
\label{fig:iters_cond_gd}
\end{figure}

This empirical observation aligns with theoretical expectations: the convergence rate of gradient descent for quadratic objectives is often characterized in terms of the condition number $\kappa(\mathbf{X}^T \mathbf{X})$ of the Hessian \cite{Nocedal2006}. A larger condition number (poorer conditioning) leads to a slower convergence rate. The pseudoinverse, being a direct analytical method, does not involve an iterative convergence process and thus its computation time is not directly dependent on the condition number in the same way, although numerical precision in calculating $(\mathbf{X}^T \mathbf{X})^{-1}$ or the SVD can be affected by extreme ill-conditioning \cite{Golub2013}.

\textbf{Key Observation on Conditioning Sensitivity:} Gradient descent's performance (both convergence speed and final accuracy) is highly sensitive to data conditioning. Poorly conditioned data poses a significant challenge for standard gradient descent, often necessitating many iterations and potentially leading to suboptimal solutions or numerical instability. Preprocessing techniques like feature scaling or regularization (e.g., Ridge regression, which adds $\lambda \mathbf{I}$ to $\mathbf{X}^T \mathbf{X}$, improving its conditioning) are often essential when using gradient-based methods \cite{Hastie2009}. The pseudoinverse method is generally more robust to ill-conditioning in terms of finding the correct minimum-norm least-squares solution, though its numerical computation can also be affected in extreme cases.

\subsection{Implications of Findings and Practical Guidance}
The comparative analysis yields several practical implications for selecting between the Moore-Penrose pseudoinverse and gradient descent for linear regression.

One key consideration is the \textbf{scale of the problem}. For datasets with a moderate number of features (e.g., $d$ up to a few hundreds or thousands, depending on $n$ and available memory) and not excessively large sample sizes, the Moore-Penrose pseudoinverse is often highly advantageous. It offers a fast, direct, and exact solution, as demonstrated by its superior runtime and consistent accuracy in my experiments (Tables \ref{tab:synthetic_results}, \ref{tab:grouped_time}, \ref{tab:grouped_err}). However, for extremely large datasets, particularly those with a very high number of features ($d \gg n$) or an extremely large number of samples (e.g., $n \gg 10^6$), the pseudoinverse's $\mathcal{O}(nd^2 + d^3)$ complexity (especially the $d^3$ term if $d$ is large, or memory constraints for storing $\mathbf{X}$) can become prohibitive. In such scenarios, iterative methods like gradient descent (especially variants like SGD or mini-batch GD) may become more practical due to their more favorable scaling per iteration, despite their iterative nature and potential need for tuning \cite{Bottou2010}.

Another critical factor is \textbf{data conditioning}. Gradient descent's pronounced vulnerability to ill-conditioned matrices (Figure \ref{fig:iters_cond_gd}, Table \ref{tab:grouped_err}) underscores the necessity of careful data preprocessing when employing this method. Techniques such as feature normalization, standardization, or dimensionality reduction (e.g., Principal Component Analysis) can help mitigate issues related to poor conditioning \cite{Bishop2006}. Regularization methods like Ridge regression (L2 regularization) are also effective as they add a penalty term that explicitly improves the conditioning of the matrix to be inverted (conceptually, $(\mathbf{X}^T \mathbf{X} + \lambda \mathbf{I})^{-1}\mathbf{X}^T \boldsymbol{y}$) \cite{Hoerl1970}. The pseudoinverse, while more robust, still benefits from well-conditioned data for numerical precision.

The possibility of \textbf{hybrid approaches} also emerges as a promising direction. One such strategy could involve using the pseudoinverse to obtain an initial solution for smaller, representative subsets of data or for a reduced-dimension version of the problem. This initial solution could then be used as a warm start for gradient descent on the full dataset, potentially accelerating its convergence and improving robustness, especially when the data conditioning is uncertain or poor. Alternatively, if $d$ is small but $n$ is massive, one could compute $\mathbf{X}^T \mathbf{X}$ (size $d \times d$) and $\mathbf{X}^T \boldsymbol{y}$ (size $d \times 1$) in a distributed or streaming fashion, then solve the small system $(\mathbf{X}^T \mathbf{X})\boldsymbol{\beta} = \mathbf{X}^T \boldsymbol{y}$ using a direct inverse if $\mathbf{X}^T \mathbf{X}$ is well-conditioned and invertible.

\subsection{Future Work and Extensions}
The findings of this study open several avenues for future research.

A natural extension is to incorporate \textbf{regularized regression techniques}. Investigating how methods like Ridge Regression, LASSO, and Elastic Net impact the performance trade-offs between direct and iterative solvers would be valuable. Regularization explicitly addresses issues of multicollinearity and overfitting, and it changes the optimization problem, potentially altering the relative strengths of the pseudoinverse (for Ridge) and gradient-based methods (necessary for LASSO due to non-differentiability of L1 norm).

Exploring \textbf{alternative and more advanced iterative methods} is another important direction. This includes comparing batch gradient descent with Stochastic Gradient Descent (SGD), mini-batch GD, and adaptive learning rate methods like AdaGrad, RMSProp, and Adam (Ruder \cite{Ruder2016}; Kingma and Ba \cite{Kingma2014Adam}). These methods are often preferred for large-scale machine learning and may exhibit different sensitivities to data characteristics. Second-order methods like Newton's method or quasi-Newton methods (e.g., L-BFGS) could also be considered, although they have higher per-iteration costs \cite{Nocedal2006}.

Further research could focus on \textbf{improving convergence and stability}. This involves a deeper study of preconditioning techniques designed to improve the conditioning of the data matrix before applying gradient descent. Methods like data whitening or more sophisticated matrix pre-conditioners could significantly mitigate GD's slow convergence in poorly conditioned settings \cite{Saad2003}.

Evaluating performance on a broader range of \textbf{real-world datasets} with diverse characteristics (e.g., sparsity, different noise distributions, varying $n/d$ ratios) is crucial to confirm whether the trends observed with synthetic data and the few benchmark datasets hold more generally.

Finally, investigating \textbf{algorithmic modifications for large-scale scenarios} is pertinent. For instance, randomized matrix approximations (e.g., sketching techniques) can be used to accelerate the computation of SVD or pseudoinverse-like solutions for very large matrices, potentially offering a compromise between the speed of iterative methods and the directness of exact solvers (Mahoney \cite{Mahoney2011}; Woodruff \cite{Woodruff2014}).

\section{Conclusion}
This study has systematically compared the Moore-Penrose pseudoinverse and batch gradient descent for solving ordinary least squares linear regression problems, focusing on computational efficiency, solution accuracy, and sensitivity to data characteristics such as dimensionality, sample size, and conditioning.

My empirical results, derived from experiments on synthetic data with controlled properties, clearly demonstrate that the \textbf{Moore-Penrose pseudoinverse is generally preferable for datasets of small to moderate size} where $n$ and $d$ are not excessively large. It provides a direct, exact solution with superior speed and consistent accuracy, irrespective of data conditioning. For instance, in my tests with up to $n=5000$ samples and $d=50$ features, the pseudoinverse computed solutions in milliseconds with low error, significantly outperforming gradient descent.

In contrast, \textbf{batch gradient descent, while a foundational iterative method, exhibits notable limitations}. Its runtime scales with both $n$ and $d$ per iteration, and the number of iterations required for convergence is highly sensitive to the conditioning of the data matrix. For poorly conditioned data, gradient descent struggled to converge to an accurate solution within a practical number of iterations and exhibited significantly higher error rates compared to the pseudoinverse. However, its iterative nature, especially when considering variants like SGD, makes it more scalable to extremely large datasets (very large $n$) where forming $\mathbf{X}^T\mathbf{X}$ or performing a full SVD for the pseudoinverse becomes computationally infeasible or memory-prohibitive.

My findings lead to practical guidelines for modelers. When dealing with datasets that fit within reasonable computational limits for direct matrix operations, the pseudoinverse is a robust and efficient choice. If dataset size (particularly $n$) is extremely large, or if computational resources are severely constrained, then iterative methods like gradient descent (or its more scalable variants) become necessary, but careful attention must be paid to hyperparameter tuning (e.g., learning rate, convergence criteria) and data preprocessing (e.g., feature scaling, regularization) to ensure stable and accurate convergence, especially if ill-conditioning is suspected.

The analysis answers my central research question by delineating these regimes: the pseudoinverse excels in moderately sized, potentially ill-conditioned problems where exactness and speed for such scales are paramount, while gradient descent (particularly its stochastic variants, not directly tested here but implied by scaling discussions) holds an advantage for massive datasets where per-iteration costs must be minimized, provided conditioning issues are managed. Future work should continue to explore these trade-offs with more advanced iterative optimizers, regularization techniques, and a wider array of real-world applications to further refine these guidelines. Hybrid approaches, leveraging the strengths of both methods, also warrant further investigation.

\section*{Code and Data Availability}
The Python code used to generate experiments and figures is available upon request.

\end{document}